\documentclass[10pt,conference,letterpaper]{IEEEtran}
\usepackage[
  letterpaper,
  left=0.75in,
  right=0.75in,
  top=1.05in,
  bottom=0.75in
]{geometry}
\usepackage{amsmath,amsfonts}
\usepackage{algorithmic}
\usepackage{array}
\usepackage[caption=false,font=normalsize,labelfont=sf,textfont=sf]{subfig}
\usepackage{textcomp}
\usepackage{stfloats}
\usepackage{url}
\usepackage{verbatim}
\usepackage{graphicx}
\usepackage{threeparttable}
\def\BibTeX{{\rm B\kern-.05em{\sc i\kern-.025em b}\kern-.08em
    T\kern-.1667em\lower.7ex\hbox{E}\kern-.125emX}}
\usepackage{balance}
\usepackage[noadjust]{cite}
\usepackage{color}
\usepackage[table, xcdraw]{xcolor}
\usepackage{comment}
\usepackage{multirow}
\usepackage{booktabs}
\usepackage{makecell}
\setcellgapes{2pt}  
\usepackage[breaklinks=true,bookmarks=false]{hyperref}
\usepackage[switch]{lineno}
\hypersetup{
colorlinks=true,
linkcolor=red
}

\definecolor{purple}{RGB}{112, 48, 160}

\IEEEoverridecommandlockouts
\begin{document}
\title{\emph{PLAF}: Pixel-wise Language-Aligned Feature Extraction for Efficient 3D Scene Understanding}

\author{Junjie Wen\textsuperscript{1}, Junlin He\textsuperscript{1,2}, Fei Ma\textsuperscript{1,3},  Jinqiang Cui\textsuperscript{1,$\dagger$} %
\thanks{This work was supported by the Major Key Project of PCL (PCL2025A03) and the National Science and Technology Major Project (2024ZD01NL00103).}
\thanks{\textsuperscript{1}Pengcheng Laboratory, Shenzhen, 518108, China}
\thanks{\textsuperscript{2}The School of Intelligent Engineering, Sun Yat-Sen University-ShenzhenCampus, Shenzhen, Guangdong, China }
\thanks{\textsuperscript{3}School of Artificial Intelligence, University of Chinese Academy of Sciences, Beijing, China.}
\thanks{$\dagger$Corresponding author: Jinqiang Cui (cuijq@pcl.ac.cn)}%
}

\maketitle

\begin{abstract}
Accurate open-vocabulary 3D scene understanding requires semantic representations that are both language-aligned and spatially precise at the pixel level, while remaining scalable when lifted to 3D space. However, existing representations struggle to jointly satisfy these requirements, and densely propagating pixel-wise semantics to 3D often results in substantial redundancy, leading to inefficient storage and querying in large-scale scenes. To address these challenges, we present \emph{PLAF}, a Pixel-wise Language-Aligned Feature extraction framework that enables dense and accurate semantic alignment in 2D without sacrificing open-vocabulary expressiveness. Building upon this representation, we further design an efficient semantic storage and querying scheme that significantly reduces redundancy across both 2D and 3D domains. Experimental results show that \emph{PLAF} provides a strong semantic foundation for accurate and efficient open-vocabulary 3D scene understanding. The codes are publicly available at \href{https://github.com/RockWenJJ/PLAF}{https://github.com/RockWenJJ/PLAF}.
\end{abstract}

\begin{IEEEkeywords}
Semantic Representation, Vision-Language Models, 3D Scene Understanding
\end{IEEEkeywords}

\section{Introduction}
Open-vocabulary 3D scene understanding seeks to recognize, localize, and reason about arbitrary concepts in real-world environments, rather than being restricted to a fixed label set~\cite{peng2023openscene}\cite{gu2024conceptgraphs}. Vision--language models (VLMs) are a natural fit for this setting because they align visual signals and natural language in a shared embedding space, enabling flexible zero-shot recognition~\cite{radford2021learning}. Despite this promise, lifting language-aligned semantics from 2D images to 3D scenes remains non-trivial, particularly when both rich semantics and precise geometric correspondence are required~\cite{jatavallabhula2023conceptfusion}\cite{takmaz2023openmask3d}.


One key reason is that commonly used vision--language features are not sufficiently fine-grained for accurate 3D semantic association. CLIP~\cite{radford2021learning}, for example, provides strong \emph{image-level} vision--language alignment, but such global semantics are often too coarse for 3D scene understanding. Several recent works~\cite{jatavallabhula2023conceptfusion}\cite{takmaz2023openmask3d} therefore attempt to map CLIP embeddings into 3D reconstructions; however, the mapped semantics can still be noisy and inconsistent in practice~\cite{ranzinger2024radio}. Meanwhile, patch-level alignment methods~\cite{alama2025rayfronts} improve spatial localization compared with image-level features, but their granularity remains limited, making them still coarse for precise 3D spatial semantic understanding. Thus, pixel-level semantic alignment is essential for precise and reliable 3D language-semantic association.

Motivated by this observation, pixel-wise language-aligned features provide a direct way to achieve such precision, offering dense semantics with fine-grained spatial correspondence at the pixel level. However, existing pixel-wise solutions face two practical challenges. First, most approaches build upon CLIP-based VLMs, yet obtaining reliable pixel-level language alignment from such models is non-trivial and often requires careful and costly feature extraction or adaptation~\cite{jatavallabhula2023conceptfusion}\cite{takmaz2023openmask3d}\cite{yu2025inst3d}. Second, even with pixel-level alignment, lifting dense features into 3D introduces substantial storage overhead: a language-aligned embedding is typically high-dimensional~\cite{radford2021learning}, and storing it for every pixel across multi-view observations quickly becomes prohibitive. These limitations call for a method that ensures accurate pixel-wise semantics and scalable efficiency in storage and querying.

To address these challenges, we present \emph{PLAF}, a \textbf{P}ixel-wise \textbf{L}anguage-\textbf{A}ligned \textbf{F}eature extraction framework for efficient open-vocabulary 3D scene understanding. To this end, we first extract dense features with a visual foundation model and aggregate them with masks from a class-agnostic mask extractor to obtain pixel-wise language-aligned features that preserve open-vocabulary expressiveness while providing fine-grained spatial correspondence in 2D. Building upon these pixel-aligned semantics, we further propose a mask-based feature storage scheme that stores compact mask-level representations instead of dense per-pixel embeddings, enabling efficient lifting to 3D as well as scalable semantic querying. Overall, this framework substantially reduces redundancy across both 2D and 3D domains, making large-scale open-vocabulary 3D semantic mapping practical.

In summary, our contributions are three-fold:
\begin{itemize}
\item We propose \emph{PLAF}, a pixel-wise language-aligned feature extraction framework that achieves dense and accurate 2D semantic alignment for open-vocabulary scene understanding.
\item We design an efficient semantic storage and querying scheme to reduce redundancy in multi-view dense semantic propagation and 3D lifting.
\item Extensive experiments demonstrate that \emph{PLAF} serves as an accurate and efficient foundation for open-vocabulary 3D scene understanding.
\end{itemize}

\section{Related Work}

\subsection{2D Foundation Models}
Recent progress in 2D foundation models has provided strong priors for open-vocabulary understanding. Vision--language models such as CLIP~\cite{radford2021learning} align images and text in a shared embedding space, while self-supervised backbones such as DINOv2~\cite{oquabdinov2} improve local feature quality and dense transferability. More recent distilled models, e.g., RADIOv2.5~\cite{heinrich2025radiov2}, further improve robustness and reduce semantic noise~\cite{ranzinger2024radio}. For precise 2D-to-3D transfer, however, image-level semantics are often too coarse, motivating methods at finer granularities, including patch-level representations~\cite{alama2025rayfronts}, region-level grouping~\cite{yu2025inst3d}, and mask-level priors such as SAM~\cite{kirillov2023segment}. Among them, mask-guided representations offer a practical balance between semantic expressiveness and spatial precision, which motivates our compact mask-level design for scalable 3D mapping.

\begin{figure*}[t!]
    \centering
    \includegraphics[width=0.9\textwidth]{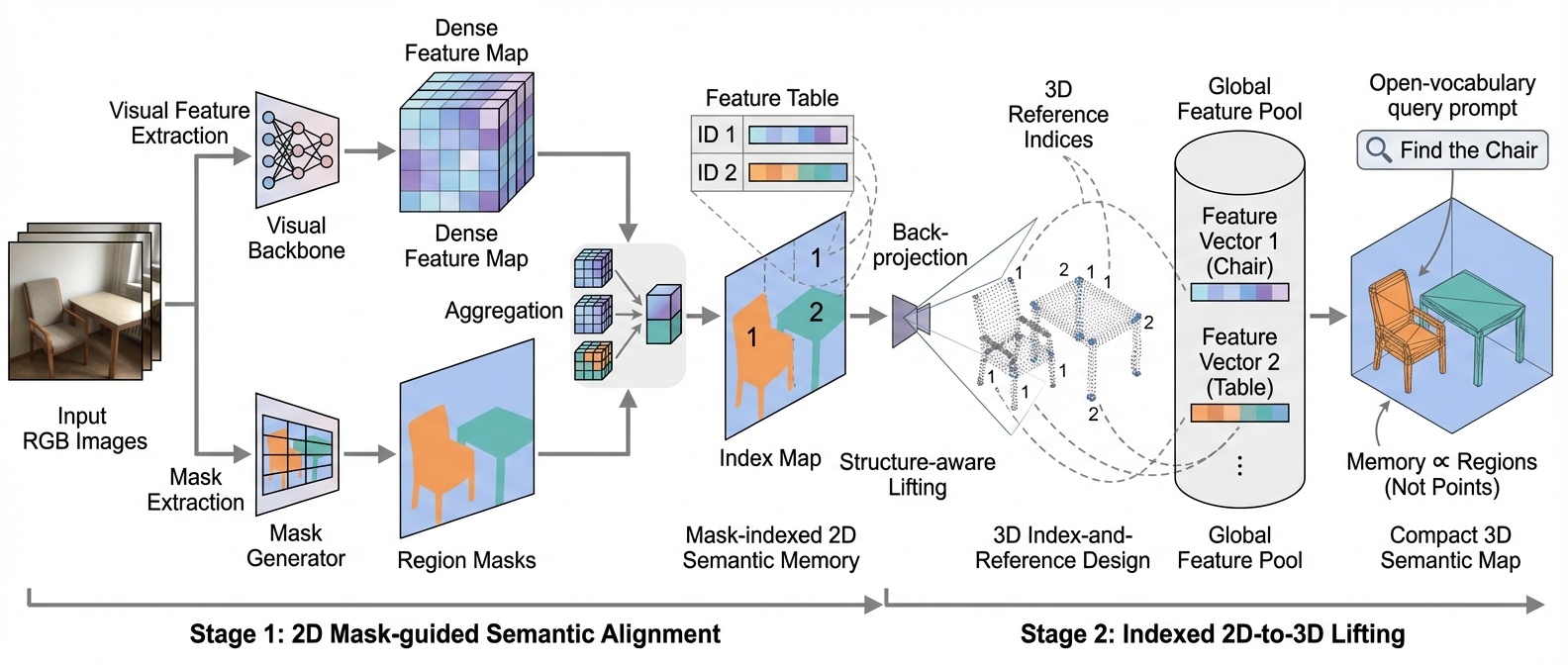}
    \caption{PLAF converts dense language-aligned pixel features into a compact mask-indexed semantic memory in 2D and extends this index-and-reference design to 3D, enabling scalable open-vocabulary 3D semantic mapping with significantly reduced redundancy.}
    \label{fig:overall_framework}
\end{figure*}

\subsection{Open-Vocabulary 3D Scene Understanding}
Open-vocabulary 3D scene understanding transfers vision--language semantics from 2D images to 3D geometry for open-set segmentation and querying~\cite{peng2023openscene}\cite{gu2024conceptgraphs}\cite{jatavallabhula2023conceptfusion}\cite{takmaz2023openmask3d}\cite{zhang2023clip}. Representative methods such as OpenScene~\cite{peng2023openscene}, ConceptFusion~\cite{jatavallabhula2023conceptfusion}, CLIP-FO3D~\cite{zhang2023clip}, OpenMask3D~\cite{takmaz2023openmask3d}, and ConceptGraphs~\cite{gu2024conceptgraphs} have shown that language-aligned 2D features can support flexible 3D understanding and interaction. However, existing methods still suffer from noisy local semantics, cross-view inconsistency, and high memory overhead when dense descriptors are attached to pixels or points. These limitations are especially severe in large indoor scenes. Our method addresses this gap by strengthening pixel-wise language alignment before 2D-to-3D lifting and improving semantic coherence in the reconstructed scene.

\subsection{Efficient 3D Spatial Semantic Representation}
Building semantic 3D maps that are both accurate and scalable requires reducing redundancy in multi-view observations while supporting fast querying. Existing open-vocabulary mapping systems often store dense per-pixel or per-point embeddings~\cite{peng2023openscene}\cite{jatavallabhula2023conceptfusion}\cite{zhang2023clip}, which is memory-intensive because high-dimensional descriptors are repeatedly observed across views. To improve efficiency, prior work has explored neural compression~\cite{huang2022neural}, sparsity-aware storage~\cite{liu2024fully}, quantization, and higher-level abstractions such as metric-semantic maps and scene graphs~\cite{gu2024conceptgraphs}\cite{rosinol2020kimera}. However, directly applying these strategies to open-vocabulary settings remains difficult, since compact representations must still preserve semantic discriminability and text-aligned geometric correspondence. Our method addresses this challenge with an index-and-reference design that is compact by construction in both 2D and 3D, substantially reducing memory usage while retaining queryability.

\section{Proposed Method}
\subsection{Overall Framework}
The overall pipeline of \emph{PLAF} is illustrated in Fig.~\ref{fig:overall_framework}. Our goal is to construct a \emph{language-grounded yet scalable} 3D semantic representation from multi-view RGB observations. To this end, \emph{PLAF} decomposes the problem into two tightly coupled components: (i) mask-guided pixel-wise alignment in 2D, which turns generic visual features into spatially grounded semantics aligned with the language space, and (ii) efficient lifting and indexing in 3D, which transforms dense 2D semantics into a compact, queryable 3D representation. In the first stage, we extract dense features with a visual foundation model and leverage masks from a class-agnostic mask extractor to aggregate features into region-consistent descriptors, improving spatial coherence while preserving open-vocabulary expressiveness. In the second stage, we replace per-pixel feature storage with a mask-indexed representation and lift these compact units into 3D, enabling efficient fusion across views and scalable semantic querying.

\subsection{Mask-guided Pixel-wise Feature Alignment}

We extract mask-guided pixel-wise features aligned with the language space from a single RGB image by combining a visual foundation model with mask priors. Existing methods often suffer from limited generalization due to task-specific supervision, rely on complex multi-stage pipelines, or trade spatial fidelity for efficiency. To address these issues, we adopt a frozen visual foundation backbone to produce stable dense features that transfer well to open-vocabulary settings, reducing semantic noise during 3D lifting~\cite{ranzinger2024radio}. We further employ a class-agnostic mask extractor to generate high-quality masks that capture object extents and fine structures. These masks enable region-consistent feature aggregation, improving spatial coherence and suppressing noise, while naturally supporting compact storage and efficient 3D lifting and querying.

\paragraph{Dense feature extraction}
Given an input image $\mathbf{I}\in\mathbb{R}^{H\times W\times 3}$, the visual foundation backbone produces a feature map $\mathbf{F}\in\mathbb{R}^{h\times w\times C}$, where $(h,w)$ is determined by the input resolution and the backbone's patch size. Since subsequent steps require pixel-wise correspondence, we upsample $\mathbf{F}$ to the original image resolution using bilinear interpolation:
\begin{equation}
\widetilde{\mathbf{F}} = \mathrm{Interp}(\mathbf{F}; H, W), \qquad
\widetilde{\mathbf{F}}\in\mathbb{R}^{H\times W\times C},
\label{eq:feature_interp}
\end{equation}
where $\widetilde{\mathbf{F}}(u,v)\in\mathbb{R}^{C}$ denotes the feature vector at pixel $(u,v)$.

\paragraph{Mask-guided feature aggregation}
Let $\{\mathbf{M}_k\}_{k=1}^{K}$ denote the binary masks predicted for $\mathbf{I}$, where $\mathbf{M}_k\in\{0,1\}^{H\times W}$. For each mask $\mathbf{M}_k$, we obtain a mask-level feature by averaging the pixel features within the mask:
\begin{equation}
\begin{split}
\mathbf{f}_k &= \frac{1}{|\Omega_k|}\sum_{(u,v)\in\Omega_k}\widetilde{\mathbf{F}}(u,v),\\
\Omega_k &= \{(u,v)\mid \mathbf{M}_k(u,v)=1\},
\end{split}
\label{eq:mask_avg}
\end{equation}
where $|\Omega_k|$ is the number of foreground pixels. In practice, this mask-guided representation reduces intra-region redundancy, improves spatial consistency, and prepares the features for efficient 2D-to-3D lifting.

\begin{figure}[t!]
    \centering
    \includegraphics[width=0.85\columnwidth]{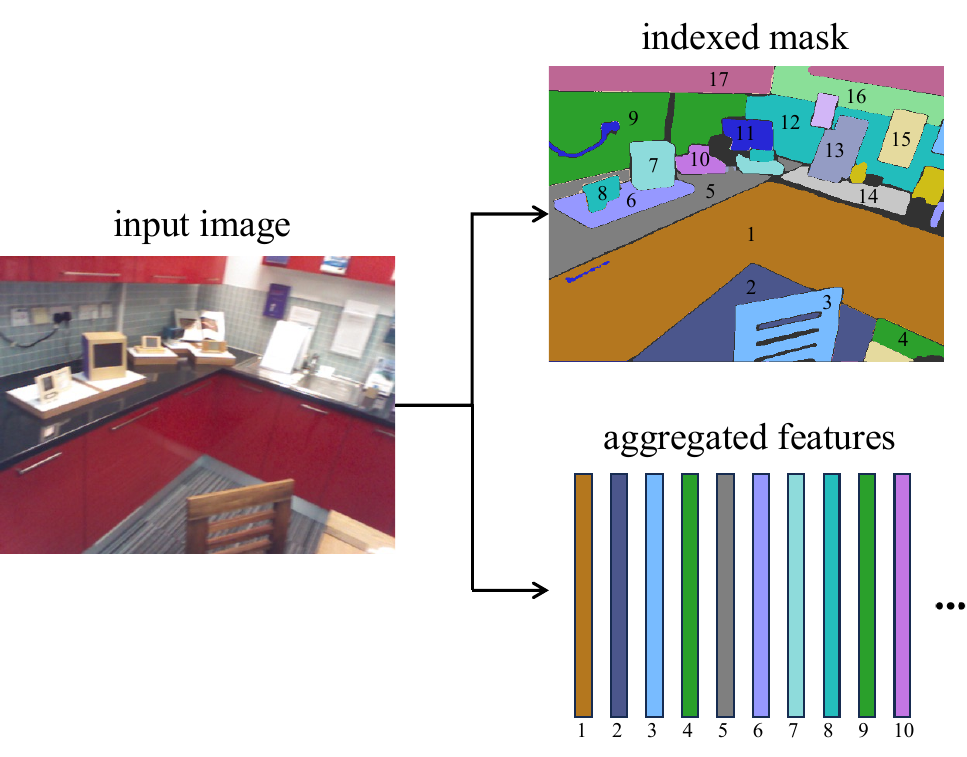}
    \caption{Mask-indexed 2D semantic memory in \emph{PLAF}. Each image is stored as an indexed mask map ($H\times W$) and a mask feature table ($K\times C$), replacing dense per-pixel feature tensors.}
    \label{fig:efficient_storage_2D}
\end{figure}

\subsection{Efficient 2D-to-3D Feature Lifting}
Given the 2D feature map and the set of masks $\{\mathbf{M}_k\}_{k=1}^{K}$, we represent each 2D observation with compact mask-level features and then lift them into 3D. Concretely, for any pixel $(u,v)$, we associate it with a mask index $\pi(u,v)$ and use the corresponding mask feature as its semantic descriptor:
\begin{equation}
\mathbf{g}(u,v) = \mathbf{f}_{\pi(u,v)}, \qquad \pi(u,v)=\arg\max_k \mathbf{M}_k(u,v),
\label{eq:pixel_assign}
\end{equation}
where $\mathbf{f}_k$ is defined in Eq.~\eqref{eq:mask_avg}. This formulation allows us to lift semantics to 3D using mask-level representations rather than storing high-dimensional features for every pixel.

\paragraph{Mask-indexed semantic memory in 2D}
Directly lifting dense per-pixel embeddings is wasteful because language-aligned features are high-dimensional and highly redundant within each mask region. We therefore store each image as a compact \emph{semantic memory} consisting of: (i) an indexed mask map $\mathbf{I}_{\text{id}}\in\mathbb{N}^{H\times W}$ that records the mask ID at each pixel, and (ii) a mask feature table $\mathbf{T}\in\mathbb{R}^{K\times C}$ that stores exactly one feature vector per mask. This decoupling turns dense pixel features into an \emph{indexing problem}: pixels only carry lightweight indices, while semantic content is stored once per region.
As illustrated in Fig.~\ref{fig:efficient_storage_2D}, this representation replaces dense per-pixel tensors with a compact index map plus a small feature table.

\begin{figure*}[t!]
    \centering
    \includegraphics[width=0.92\textwidth]{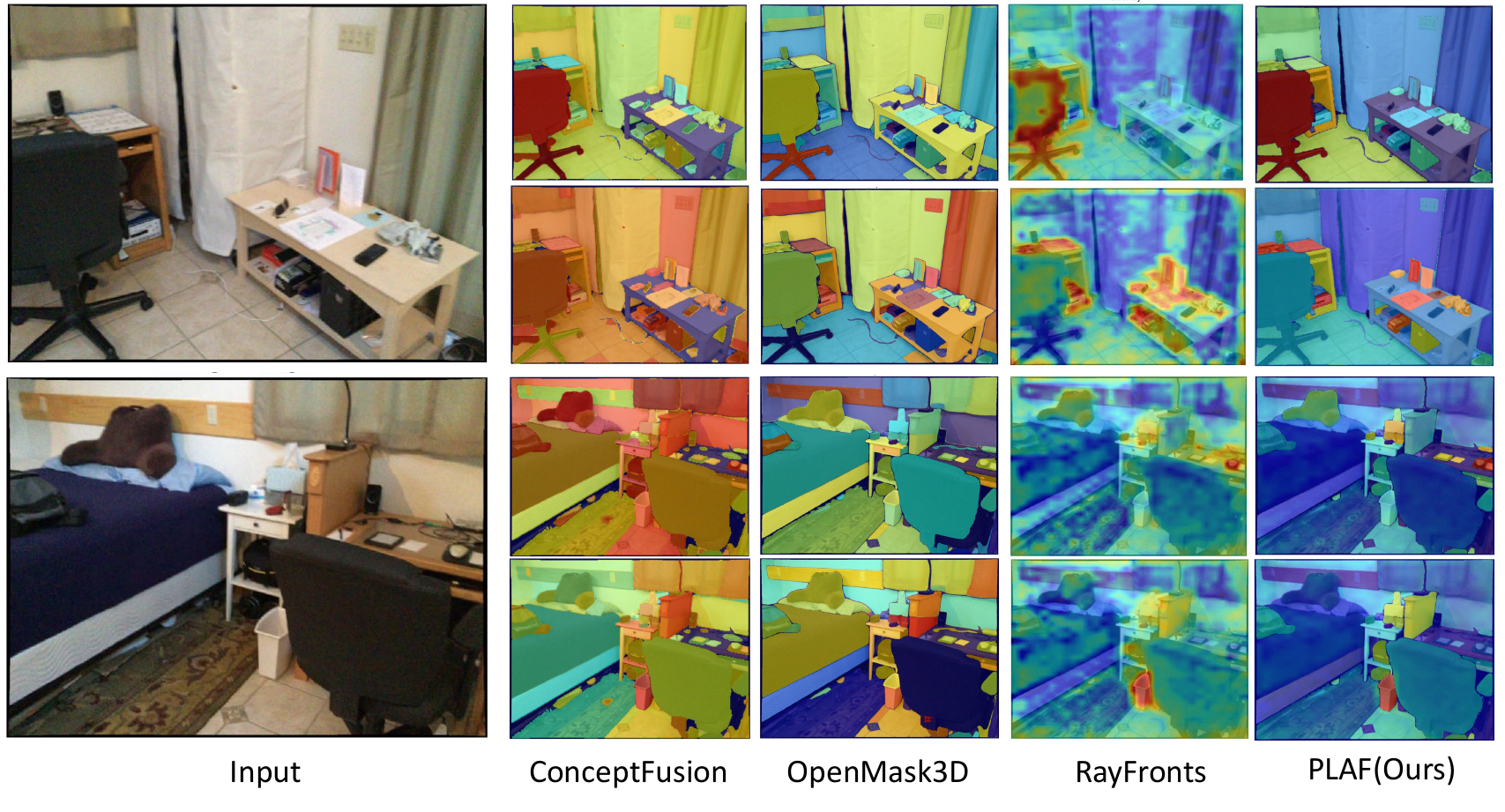}
    \caption{Qualitative comparison of 2D text-query results on ScanNet. The leftmost column shows input RGB images; from left to right, the remaining columns show ConceptFusion, OpenMask3D, RayFronts~\cite{alama2025rayfronts}, and \emph{PLAF} (ours). From top to bottom, each row corresponds to the text queries \emph{chair}, \emph{book}, \emph{mouse}, and \emph{dustbin}. Although ConceptFusion and OpenMask3D produce reasonable segmentation maps, they often mis-localize queried semantics. RayFronts captures coarse semantic regions but remains spatially imprecise due to patch-wise semantics. In contrast, \emph{PLAF} yields more accurate and sharper localization for open-vocabulary objects.}
    \label{fig:text_query}
\end{figure*}

Assume the feature dimension is $C$, each floating-point element takes $b_f$ bytes (e.g., $4$ for FP32), and each mask index takes $b_i$ bytes (e.g., $2$ for uint16). \text{Traditional dense storage} keeps one $C$-dimensional feature for every pixel, whose cost is:
\begin{equation}
\mathcal{S}_{\text{dense}}^{\text{2D}} = HW \cdot C \cdot b_f.
\label{eq:storage_dense}
\end{equation}
In contrast, our \text{proposed mask-indexed storage} keeps (i) an indexed mask map (one integer ID per pixel) and (ii) a mask feature table (one $C$-dimensional feature per mask). Let $K$ denote the number of masks in this image. The total cost is:
\begin{equation}
\mathcal{S}_{\text{mask}}^{\text{2D}} = HW \cdot b_i + K \cdot C \cdot b_f,
\label{eq:storage_mask}
\end{equation}
The corresponding reduction ratio is
\begin{equation}
\rho = \frac{\mathcal{S}_{\text{mask}}}{\mathcal{S}_{\text{dense}}}
= \frac{b_i}{C b_f} + \frac{K}{HW}.
\label{eq:storage_ratio}
\end{equation}
Since typically $K\ll HW$ and $C$ is large, the first term is small and the second term decreases with image resolution.

For an image of size $H\times W=480\times 640$ with $C=1024$, using FP32 features ($b_f=4$) and uint16 indices ($b_i=2$), dense storage requires $480\cdot 640\cdot 1024\cdot 4\approx 1.26$~GB per image. In contrast, with $K=200$ masks, our mask-indexed scheme requires $480\cdot 640\cdot 2 + 200\cdot 1024\cdot 4\approx 1.43$~MB, which is about $0.11\%$ of dense storage.

\paragraph{Efficient lifting and fusion in 3D}
Building upon the 2D semantic memory, we lift semantics to 3D by back-projecting pixels to 3D and attaching the corresponding mask-level descriptor $\mathbf{f}_{\pi(u,v)}$ (Eq.~\eqref{eq:pixel_assign}) to each observation. We then fuse observations across views by aggregating mask-level descriptors into a compact set of 3D point clouds that reference a small pool of semantic descriptors. Compared with storing a language-aligned feature for every 3D point, this \emph{index-and-reference} design reduces redundancy across both multi-view images and the reconstructed 3D map, while supporting scalable semantic querying.

For a point-cloud map with $N$ 3D points and feature dimension $C$, the naive per-point semantic storage cost is
\begin{equation}
\mathcal{S}^{3\text{D}}_{\text{dense}} = N \cdot C \cdot b_f.
\label{eq:storage_3d_dense}
\end{equation}
In our index-and-reference design, we maintain (i) a compact feature pool of size $M$ (i.e., $M$ unique semantic descriptors after fusion) and (ii) an integer reference for each 3D point that points to an entry in the pool. Let $b_r$ denote the reference size in bytes (e.g., $2$ or $4$). The resulting storage cost is
\begin{equation}
\mathcal{S}^{3\text{D}}_{\text{index}} = N \cdot b_r + M \cdot C \cdot b_f,
\label{eq:storage_3d_index}
\end{equation}
which yields the ratio
\begin{equation}
\rho_{3\text{D}} = \frac{\mathcal{S}^{3\text{D}}_{\text{index}}}{\mathcal{S}^{3\text{D}}_{\text{dense}}}
= \frac{b_r}{C b_f} + \frac{M}{N}.
\label{eq:storage_3d_ratio}
\end{equation}
When $M\ll N$, this provides substantial memory reduction while retaining queryability.

\begin{figure*}[t!]
    \centering
    \includegraphics[width=0.92\textwidth]{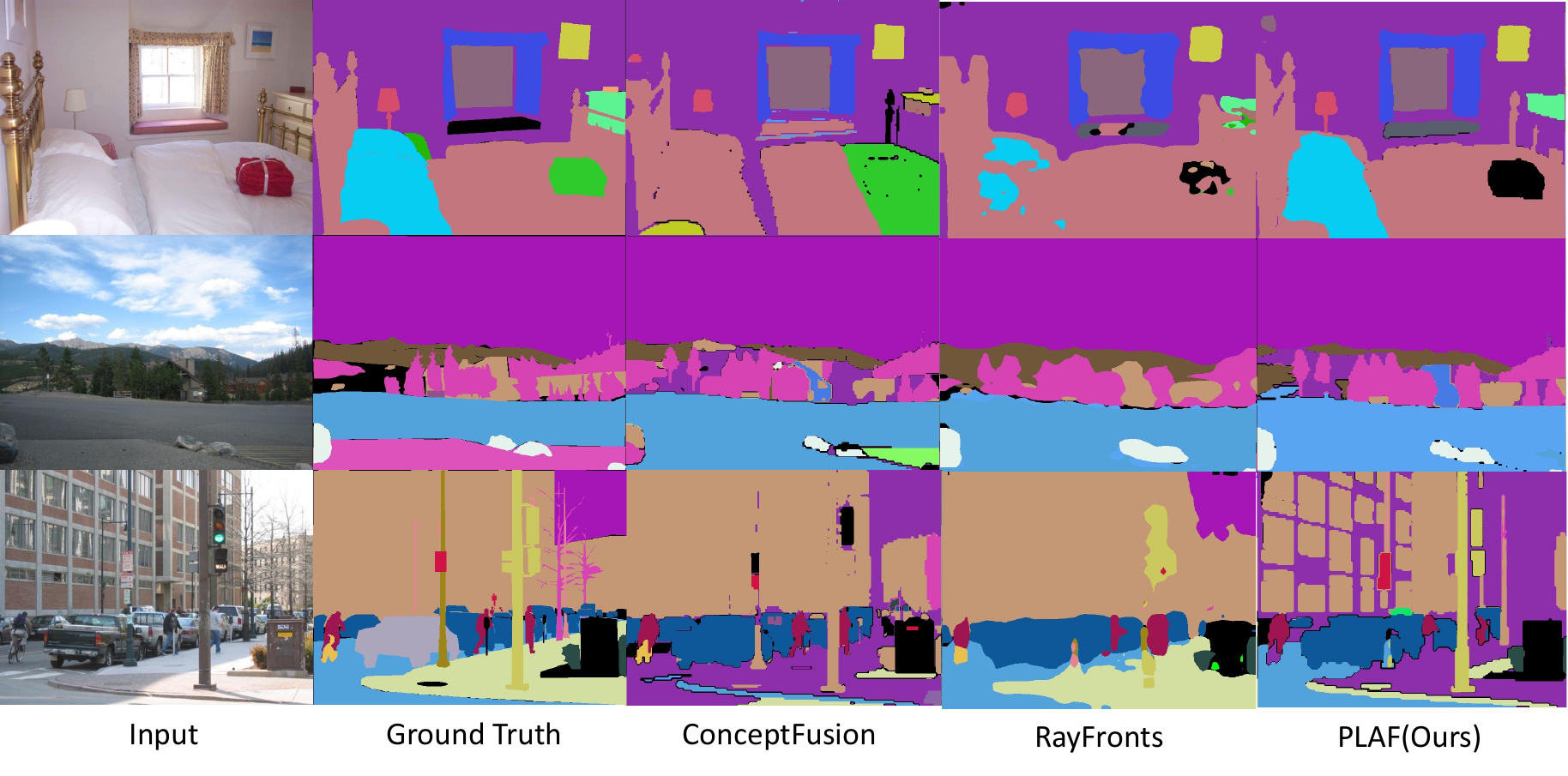}
    \caption{Qualitative linear-probe segmentation results on ADE20K. From left to right: input image, ground truth, ConceptFusion, RayFronts, and \emph{PLAF} (ours). Compared with ConceptFusion and RayFronts, \emph{PLAF} produces masks that are better aligned with object/region boundaries and shows fewer semantic confusions in complex scenes.}
    \label{fig:linear_probe_2d}
\end{figure*}

As an illustrative example, let $N=10^7$ points, $C=1024$, FP32 features ($b_f=4$), and uint16 references ($b_r=2$). Traditional dense per-point storage costs $10^7\cdot 1024\cdot 4\approx 40.96$~GB for semantics alone, which is typically impractical for real-world large-scale mapping (especially when multiple scenes or time steps must be retained). With a feature pool of size $M=10^4$, our storage costs $10^7\cdot 2 + 10^4\cdot 1024\cdot 4\approx 60.96$~MB, which is about $0.14\%$ of dense storage and makes large-scale deployment feasible.

\section{Experiments}
\subsection{Experimental Setup}
We evaluate \emph{PLAF} on ScanNet~\cite{dai2017scannet} under a zero-shot setting, where no task-specific training is performed and predictions are generated from frozen feature backbones and their lifted representations. We use RADIOv2.5~\cite{heinrich2025radiov2} as the feature extractor and SAM~\cite{kirillov2023segment} as the mask extractor. For fair comparison, baselines (ConceptFusion~\cite{jatavallabhula2023conceptfusion} and OpenMask3D~\cite{takmaz2023openmask3d}) are also evaluated in zero-shot mode on the same ScanNet test scenes with identical text prompts.

\subsection{2D Text Query on ScanNet}

Fig.~\ref{fig:text_query} compares 2D text-query results under four open-vocabulary queries (\emph{chair}, \emph{book}, \emph{mouse}, and \emph{dustbin}) across two ScanNet scenes. ConceptFusion~\cite{jatavallabhula2023conceptfusion} and OpenMask3D~\cite{takmaz2023openmask3d} generally provide plausible segmentation maps, but their semantic localization is often inaccurate, especially when queried objects are small, partially occluded, or close to visually similar regions. RayFronts~\cite{alama2025rayfronts} can roughly indicate relevant semantic regions, yet its patch-wise representation leads to coarse responses and imprecise object boundaries. By contrast, \emph{PLAF} provides tighter object alignment and cleaner boundaries, and it suppresses off-target activations in nearby distractor regions. These qualitative results indicate that pixel-wise language-aligned features are more reliable for fine-grained open-vocabulary localization in cluttered indoor scenes.

\subsection{Semantic Segmentation on ADE20K with Linear Probe}
To further validate the representation quality of our extracted features, we conduct a linear-probe semantic segmentation experiment on ADE20K~\cite{zhou2017scene}. This setting isolates the intrinsic expressiveness of frozen features by training only a lightweight linear classifier, so better performance indicates stronger semantic separability and transferability.

As shown in Fig.~\ref{fig:linear_probe_2d}, the predictions from ConceptFusion and RayFronts contain more boundary leakage and misclassified regions in both indoor and outdoor scenes. In contrast, \emph{PLAF} yields cleaner boundaries and more stable class assignments. These qualitative observations are consistent with Table~\ref{tab:ade20k_linear_probe}, where \emph{PLAF} achieves the best mIoU of 41.1\%, outperforming RayFronts (39.5\%) by 1.6 points and ConceptFusion (27.1\%) by 14.0 points.

\begin{table}[t]
    \centering
    \caption{Linear-probe semantic segmentation on ADE20K (mIoU, higher is better).}
    \label{tab:ade20k_linear_probe}
    \begin{tabular}{lccc}
        \toprule
        Method & ConceptFusion & RayFronts & \emph{PLAF} (Ours) \\
        \midrule
        mIoU ($\uparrow$) & 27.1\% & 39.5\% & \textbf{41.1\%} \\
        \bottomrule
    \end{tabular}
\end{table}

\subsection{3D Text Query on ScanNet}
\begin{figure*}[t]
    \centering
    \includegraphics[width=0.92\textwidth]{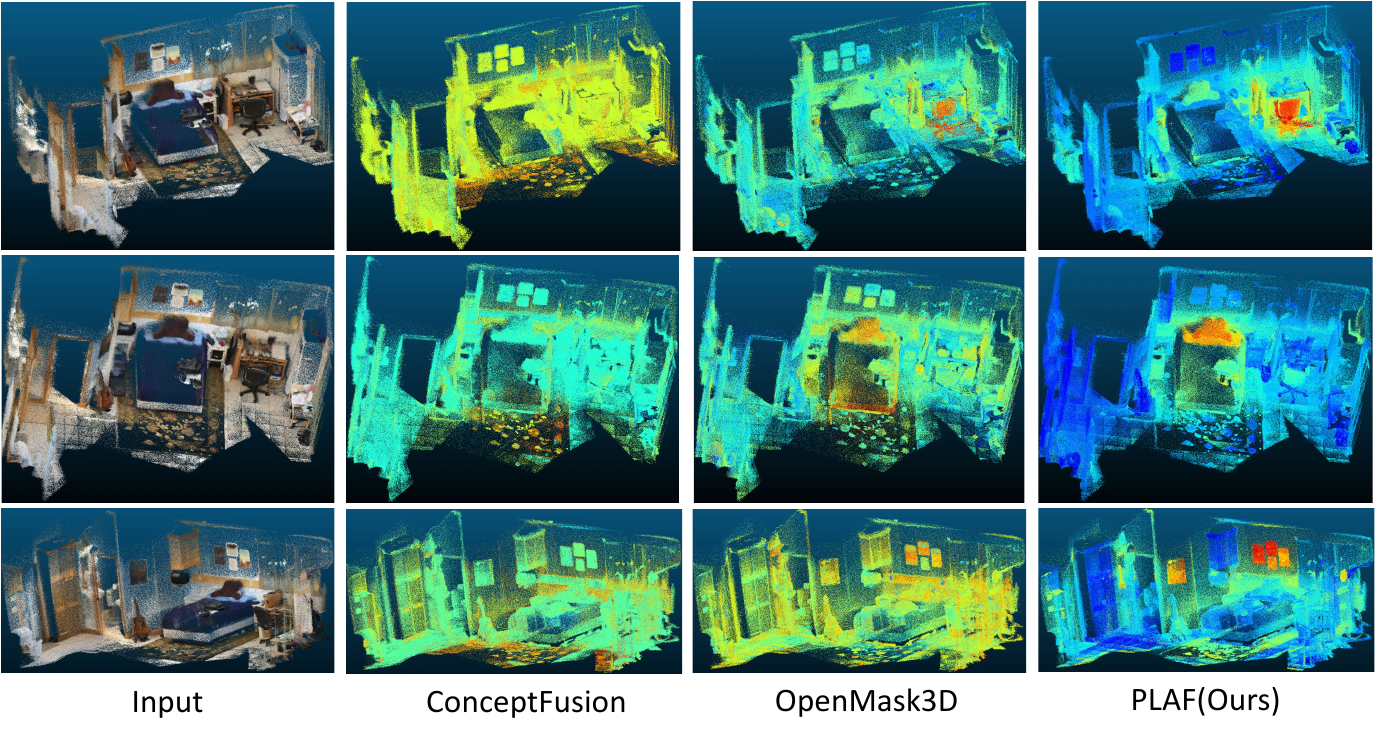}
    \caption{Qualitative comparison of 3D text-query results on ScanNet. From top to bottom, the text queries are \emph{chair}, \emph{pillow}, and \emph{picture}. From left to right, columns show the input RGB point cloud, ConceptFusion~\cite{jatavallabhula2023conceptfusion}, OpenMask3D~\cite{takmaz2023openmask3d}, and \emph{PLAF} (ours). Compared with ConceptFusion and OpenMask3D, \emph{PLAF} provides more accurate and clearer localization of queried objects, with sharper boundaries and fewer off-target responses.}
    \label{fig:text_query_3d}
\end{figure*}

We conduct a qualitative comparison for open-vocabulary 3D text queries on ScanNet, focusing on ConceptFusion~\cite{jatavallabhula2023conceptfusion}, OpenMask3D~\cite{takmaz2023openmask3d}, and our \emph{PLAF}. For each method, text embeddings are matched with lifted 3D semantic descriptors, and the query responses are visualized in the same scene.

As shown in Fig.~\ref{fig:text_query_3d}, the three queries (from top to bottom) are \emph{chair}, \emph{pillow}, and \emph{picture}, and the compared results (from left to right) are the input RGB point cloud, ConceptFusion, OpenMask3D, and \emph{PLAF}. ConceptFusion and OpenMask3D tend to produce diffuse activations and ambiguous localization around the targets. In contrast, \emph{PLAF} consistently concentrates responses on the queried objects and yields clearer boundaries, demonstrating more accurate and reliable 3D open-vocabulary localization.

\section{Conclusion}
In this paper, we presented \emph{PLAF}, a pixel-wise language-aligned feature extraction framework for efficient open-vocabulary 3D scene understanding. By combining dense foundation-model features with class-agnostic mask priors and an index-and-reference storage design, \emph{PLAF} improves fine-grained semantic localization in both 2D and 3D while drastically reducing representation redundancy. Across ScanNet qualitative text-query results and ADE20K linear-probe evaluation, our method demonstrates stronger localization quality and semantic separability than representative baselines, while reducing semantic storage by over 99\% relative to dense per-pixel/per-point schemes, indicating strong potential for scalable real-world 3D open-vocabulary mapping.

\bibliographystyle{IEEEtran}
\bibliography{references}

@inproceedings{radford2021learning,
  title={Learning transferable visual models from natural language supervision},
  author={Radford, Alec and Kim, Jong Wook and Hallacy, Chris and Ramesh, Aditya and Goh, Gabriel and Agarwal, Sandhini and Sastry, Girish and Askell, Amanda and Mishkin, Pamela and Clark, Jack and others},
  booktitle={International conference on machine learning},
  pages={8748--8763},
  year={2021},
  organization={PMLR}
}

@inproceedings{peng2023openscene,
  title={OpenScene: 3D scene understanding with open vocabularies},
  author={Peng, Songyou and Genova, Kyle and Jiang, Chiyu and Tagliasacchi, Andrea and Pollefeys, Marc and Funkhouser, Thomas},
  booktitle={Proceedings of the IEEE/CVF Conference on Computer Vision and Pattern Recognition},
  pages={815--824},
  year={2023}
}

@inproceedings{jatavallabhula2023conceptfusion,
  title={ConceptFusion: Open-set multimodal 3D mapping},
  author={Jatavallabhula, Krishna Murthy and Kuwajerwala, Alihusein and Gu, Qiao and Omama, Mohd and Chen, Tao and Maalouf, Alaa and Li, Shuang and Iyer, Ganesh and Saryazdi, Soroush and Keetha, Nikhil and others},
  booktitle={2023 IEEE International Conference on Robotics and Automation (ICRA)},
  pages={11508--11514},
  year={2023},
  organization={IEEE}
}

@inproceedings{zhang2023clip,
  title={CLIP-FO3D: Learning free open-world 3D scene representations from 2D dense CLIP},
  author={Zhang, Junbo and Dong, Runpei and Ma, Kaisheng},
  booktitle={Proceedings of the IEEE/CVF International Conference on Computer Vision},
  pages={17288--17299},
  year={2023}
}

@inproceedings{huang2022neural,
  title={Neural compression-based feature learning for video restoration},
  author={Huang, Cong and Li, Jiahao and Li, Bin and Liu, Dong and Lu, Yan},
  booktitle={Proceedings of the IEEE/CVF Conference on Computer Vision and Pattern Recognition},
  pages={5872--5881},
  year={2022}
}

@inproceedings{liu2024fully,
  title={Fully sparse 3D occupancy prediction},
  author={Liu, Haisong and Chen, Yang and Wang, Haiguang and Yang, Zetong and Li, Tianyu and Zeng, Jia and Chen, Li and Li, Hongyang and Wang, Limin},
  booktitle={Proceedings of the IEEE/CVF Conference on Computer Vision and Pattern Recognition},
  pages={17500--17510},
  year={2024}
}

@inproceedings{rosinol2020kimera,
  title={Kimera: an open-source library for real-time metric-semantic localization and mapping},
  author={Rosinol, Antoni and Abate, Marcus and Chang, Yun and Carlone, Luca},
  booktitle={2020 IEEE International Conference on Robotics and Automation (ICRA)},
  pages={1689--1696},
  year={2020},
  organization={IEEE}
}

@inproceedings{gu2024conceptgraphs,
  title={Conceptgraphs: Open-vocabulary 3d scene graphs for perception and planning},
  author={Gu, Qiao and Kuwajerwala, Ali and Morin, Sacha and Jatavallabhula, Krishna Murthy and Sen, Bipasha and Agarwal, Aditya and Rivera, Corban and Paul, William and Ellis, Kirsty and Chellappa, Rama and others},
  booktitle={2024 IEEE International Conference on Robotics and Automation (ICRA)},
  pages={5021--5028},
  year={2024},
  organization={IEEE}
}

@article{takmaz2023openmask3d,
  title={OpenMask3D: Open-Vocabulary 3D Instance Segmentation},
  author={Takmaz, Ayca and Fedele, Elisabetta and Sumner, Robert and Pollefeys, Marc and Tombari, Federico and Engelmann, Francis},
  journal={Advances in Neural Information Processing Systems},
  volume={36},
  pages={68367--68390},
  year={2023}
}

@inproceedings{ranzinger2024radio,
  title={Am-radio: Agglomerative vision foundation model reduce all domains into one},
  author={Ranzinger, Mike and Heinrich, Greg and Kautz, Jan and Molchanov, Pavlo},
  booktitle={Proceedings of the IEEE/CVF conference on computer vision and pattern recognition},
  pages={12490--12500},
  year={2024}
}

@article{alama2025rayfronts,
  title={RayFronts: Open-Set Semantic Ray Frontiers for Online Scene Understanding and Exploration},
  author={Alama, Omar and Bhattacharya, Avigyan and He, Haoyang and Kim, Seungchan and Qiu, Yuheng and Wang, Wenshan and Ho, Cherie and Keetha, Nikhil and Scherer, Sebastian},
  journal={arXiv preprint arXiv:2504.06994},
  year={2025}
}

@inproceedings{yu2025inst3d,
  title={Inst3d-lmm: Instance-aware 3d scene understanding with multi-modal instruction tuning},
  author={Yu, Hanxun and Li, Wentong and Wang, Song and Chen, Junbo and Zhu, Jianke},
  booktitle={Proceedings of the Computer Vision and Pattern Recognition Conference},
  pages={14147--14157},
  year={2025}
}

@inproceedings{heinrich2025radiov2,
  title={Radiov2. 5: Improved baselines for agglomerative vision foundation models},
  author={Heinrich, Greg and Ranzinger, Mike and Yin, Hongxu and Lu, Yao and Kautz, Jan and Tao, Andrew and Catanzaro, Bryan and Molchanov, Pavlo},
  booktitle={Proceedings of the Computer Vision and Pattern Recognition Conference},
  pages={22487--22497},
  year={2025}
}

@inproceedings{kirillov2023segment,
  title={Segment anything},
  author={Kirillov, Alexander and Mintun, Eric and Ravi, Nikhila and Mao, Hanzi and Rolland, Chloe and Gustafson, Laura and Xiao, Tete and Whitehead, Spencer and Berg, Alexander C and Lo, Wan-Yen and others},
  booktitle={Proceedings of the IEEE/CVF international conference on computer vision},
  pages={4015--4026},
  year={2023}
}

@article{oquabdinov2,
  title={DINOv2: Learning Robust Visual Features without Supervision},
  author={Oquab, Maxime and Darcet, Timoth{\'e}e and Moutakanni, Th{\'e}o and Vo, Huy V and Szafraniec, Marc and Khalidov, Vasil and Fernandez, Pierre and HAZIZA, Daniel and Massa, Francisco and El-Nouby, Alaaeldin and others},
  journal={Transactions on Machine Learning Research},
  year={2023}
}

@inproceedings{dai2017scannet,
    title={ScanNet: Richly-annotated 3D Reconstructions of Indoor Scenes},
    author={Dai, Angela and Chang, Angel X. and Savva, Manolis and Halber, Maciej and Funkhouser, Thomas and Nie{\ss}ner, Matthias},
    booktitle = {Proc. Computer Vision and Pattern Recognition (CVPR), IEEE},
    year = {2017}
}

@inproceedings{zhou2017scene,
  title={Scene parsing through ade20k dataset},
  author={Zhou, Bolei and Zhao, Hang and Puig, Xavier and Fidler, Sanja and Barriuso, Adela and Torralba, Antonio},
  booktitle={Proceedings of the IEEE conference on computer vision and pattern recognition},
  pages={633--641},
  year={2017}
}

\end{document}